\documentclass{article}

\usepackage{chngpage}

\usepackage[final]{nips_2017}

\usepackage[utf8]{inputenc} 
\usepackage[T1]{fontenc}    
\usepackage{hyperref}       
\usepackage{url}            
\usepackage{booktabs}       
\usepackage{amsfonts}       
\usepackage{nicefrac}       
\usepackage{microtype}      

\title{A global feature extraction model for the effective computer aided diagnosis of mild cognitive impairment using structural MRI images}

\author{
  Chen Fang, Panuwat Janwattanapong, Chunfei Li, and Malek Adjouadi \\
  Center for Advanced Technology and Education\\
  Florida International University\\
  Miami, FL 33174 \\
  \texttt{chfang@fiu.edu} \\
  \And
  For the Alzheimer’s Disease Neuroimaging Initiative \\
}

\begin{document}

\maketitle

\begin{abstract}
  Multiple modalities of biomarkers have been proved to be very sensitive in assessing the progression of Alzheimer's disease (AD), and using these modalities and machine learning algorithms, several approaches have been proposed to assist in the early diagnosis of AD. Among the recent investigated state-of-the-art approaches, Gaussian discriminant analysis (GDA)-based approaches have been demonstrated to be more effective and accurate in the classification of AD, especially for delineating its prodromal stage of mild cognitive impairment (MCI). Moreover, among those binary classification investigations, the local feature extraction methods were mostly used, which made them hardly be applied to a practical computer aided diagnosis system. Therefore, this study presents a novel global feature extraction model taking advantage of the recent proposed GDA-based dual high-dimensional decision spaces, which can significantly improve the early diagnosis performance comparing to those local feature extraction methods. In the true test using $20\%$ held-out data, for discriminating the most challenging MCI group from the cognitively normal control (CN) group, an F1 score of $91.06\%$, an accuracy of $88.78\%$, a sensitivity of $91.80\%$, and a specificity of $83.78\%$ were achieved that can be considered as the best performance obtained so far. 
\end{abstract}

\section{Introduction}

In order to facilitate the planning of early intervention and treatment in Alzheimer’s disease (AD), one of the leading cause of death for older people in the United States, over the past few years, several machine learning-based approaches have been proposed to assist in the early computer aided diagnosis (CAD) of AD [1]. Using multiple modalities of biomarkers that have been proved to be sensitive in assessing the progression of AD, those proposed approaches aimed at the effective and accurate classification performance, especially for discriminating the subtler prodromal stage of AD, mild cognitive impairment (MCI) group, from the cognitively normal control (CN) group. Among those modalities, structural magnetic resonance imaging (MRI) is currently widely used for analyzing the gradual progression of atrophy patterns in key brain regions [1]. And utilizing structural MRI images, the recent proposed Gaussian discriminant analysis (GDA)-based approaches have been demonstrated to be effective and accurate in the classification of AD and MCI [2]. Moreover, since most of the current studies assumed only binary classification, where classification experiments were based on two-group comparisons, i.e., CN vs. MCI, CN vs. AD, and MCI vs. AD, such studies commonly applied local feature extraction methods. For different comparisons, the feature sets in the classification are very various from each other, which also limit the clinical diagnosis for a given patient, which could belong in any of the three groups [2]. 

This study aims to introduce a novel global feature extraction model based on the recent proposed GDA-based dual high-dimensional decision spaces, so that can significantly improve the early CAD performance. Instead of extracting three different sets of features for those three comparisons (CN vs. MCI, CN vs. AD, and MCI vs. AD), respectively, in this study, only one optimal set of features will be generated. The optimal set of features in this study indicates that the entorhinal cortex is the most significant cortical region associated with the progression of AD, which is consistent with recent studies concluding that the entorhinal cortex is the signature region to be implicated in AD [3]–[5]. And the top 5 regions of interest, entorhinal, middle temporal, inferior temporal, fusiform, and parahippocampal are all AD signature regions investigated previously [6]. Furthermore, for the dual high-dimensional decision spaces, a global GDA-based classifier used for combined training will be proposed as well and achieve a significant improvement of the early CAD performance. 

\section{Material}

\subsection{Subjects}

The data used in the preparation of this study were obtained from the Alzheimer’s disease Neuroimaging Initiative (ADNI) database (adni.loni.usc.edu), as part of the ADNI1: Complete 1Yr 1.5T collection and their assessments at baseline, which includes 628 individuals (190 CN, 305 MCI, and 133 AD) [7]. The ADNI was launched in 2003 as a public-private partnership, led by Principal Investigator Michael W. Weiner, MD. The primary goal of ADNI has been to test whether serial magnetic resonance imaging (MRI), positron emission tomography (PET), other biological markers, and clinical and neuropsychological assessment can be combined to measure the progression of MCI and early AD. The primary phenotype is diagnostic group and MMSE. All source imaging data consist of 1.5 Tesla T1-weighted MRI volumes in the NIfTI (.nii.gz) format from the ADNI1: Complete 1Yr 1.5T Data Collection. 

\subsection{MRI data pre-processing}

Using three neuroimaging software pipelines: FreeSurfer [8], Advanced Normalization Tools (ANTs) [9], and Mindboggle [10], the original MRI data were pre-processed following the instruction provided by Alzheimer's Disease Big Data DREAM Challenge \#1 [11]. Tables of morphometric data were derived from the images using the following seven shape measures for all 25 FreeSurfer labeled cortical regions for both left and right hemispheres of the brain: 1) surface area; 2) travel depth; 3) geodesic depth; 4) mean curvature; 5) convexity; 6) thickness; 7) volume. FreeSurfer pipeline (version 5.3) was applied to all T1-weighted images to generate labeled cortical surfaces, and labeled cortical and noncortical volumes. Templates and atlases used by ANTs and Mindboggle could be found on the Mindboggle website [12]. The aforementioned pre-processed MRI data of the 25 labeled cortical regions were used to generate two 175-variable ($7\times25$) vector discriminator, for each subject (one 175-variable vector per hemisphere). Any subjects that involved abnormal variables were eliminated from the dataset for further investigation, for example, some of their cortical regions having measurements of some features to be zero. 

\section{Methods}

\subsection{ANOVA-based feature ranking}

The analysis of variance (ANOVA) was carried out on each of the 175 features of the two brain hemispheres between all three groups to determine the significance of each feature in terms of classification outcome and all features were thereafter ranked according to their p-values. For the purpose of comparing the performance between the local and global feature extraction models, ANOVA was also performed on each of the 175 features of the two brain hemispheres between any two groups (i.e., CN vs. MCI, CN vs. AD, and MCI vs. AD), which would show that the significant features of each comparison are much different to others in Section \ref{res} .  

\subsection{Incremental error analysis}

In order to maintain only few but key features for achieving best classification performance, an incremental error analysis was employed to determine how many of the top-ranked features ought to be involved in the final GDA-based classifier. In the initial phase, the proposed GDA-based classifier only used the first-ranked feature. The error analysis was performed whereby introducing the next top-ranked feature in the classifier at each subsequent phase, and recording the corresponding classification statistics, which then would be compared with the previous phase. In this study, as a machine learning classification problem, the performance was estimated by the F1 score, and until the performance can be no longer improved, the optimal sets of features should have been obtained. 

\subsection{Global GDA-based classifier}

Taking advantage of GDA, an important generative learning algorithm for high dimensional classification problems, some proposed studies were capable of recognizing the different patterns between any given groups (i.e., CN vs. MCI, CN vs. AD, and MCI vs. AD) [2]. The fundamental part of GDA is the multivariate Gaussian distribution or multivariate normal distribution, which could take the correlations among all features into account of the proposed global classifier. The performance of the proposed global GDA-based model was measured by some classification related statistics based on tenfold cross validation using $80\%$ of the noise-free detected data points, including the F1 score, accuracy (ACC), sensitivity (SEN), specificity (SPE) and so on. After the optimal sets of features were obtained, the remaining $20\%$ of the noise-free detected data were utilized as held-out test data to estimate the classification performance.

As the same as aforementioned ANOVA-based feature ranking process, both local and global classifiers were validated. For the local model, the GDA-based classifier was applied to each hemisphere of the brain separately, and corresponding to each hemisphere, an optimal set of features was generated, and the two sets were then used as the final optimal sets of features. In contrast with the local model, the proposed global model implemented a combined incremental error analysis on the 175 features of the left and right hemispheres of the brain, so that all combinations of the two sets of 175 features were taken into consideration for obtaining the final optimal sets of features. In terms of the dual decision space CAD, if either one of the two decision spaces produced a positive result, the corresponding subject should be classified as such. 

\section{Results and conclusion}
\label{res}

\subsection{Top-ranked features}

\begin{table}[t]
	\caption{Number of significant features selected for each model}
	\label{table1}
	\centering
	\begin{tabular}{lllll}
		\toprule
		Model    
		& \multicolumn{3}{c}{Local}  & Global      \\
		\cmidrule{2-4}
		Comparison & CN vs. MCI      & CN vs. AD     & MCI vs. AD    & All \\
		\cmidrule{2-5}
		Side of brain 
		& \multicolumn{4}{c}{Number of significant variables (\textit{p-value} < 0.01)} \\
		\midrule
		Left       & 50              & 79            & 51            & 71  \\
		Right      & 44              & 68            & 41            & 66  \\
		\bottomrule
	\end{tabular}
\end{table}

After pre-processing the MRI images, 9 subjects were eliminated because of the noisy data which included measurements with zero values, so the final data used in the classification experiment included 619 individuals, among them 187 CN, 301 MCI, and 131 AD. As mentioned earlier, in the local feature ranking, ANOVA was performed for CN vs. MCI, CN vs. AD, and MCI vs. AD using two 175-feature vectors corresponding to the left and right hemispheres of the brain. For each comparison, all features found at 0.01 level of significance (LOS) out of all 175 features for each side of the brain were used for the tenfold cross validation. In the global case, ANOVA was carried out on the two sets of 175-feature vectors between all three groups. For each model, the number of selected features is shown in Table \ref{table1}. 

In the local feature ranking, for each comparison, the top ranked feature sets are much different to others. Nevertheless, in both local and global feature ranking results, it could be observed that the most significant cortical region associated with the progression of AD is the entorhinal cortex, that is consistent with recent studies indicating that indeed the entorhinal cortex is the first area to be implicated in AD [3]–[5]. Furthermore, the entorhinal cortex has been also proven as a major source of projections to the hippocampus [13]. The other four significant regions of interest identified from the global feature ranking, middle temporal, inferior temporal, fusiform, and parahippocampal are all AD signature regions investigated previously [6], which provide credence to the validity of our ANOVA-based feature ranking method. 

\subsection{Optimal feature sets}

The summary of tenfold cross validation performance is demonstrated in Table \ref{table2}. For both models, it can be observed that the final optimal sets obtained by the global GDA-based classifiers are different from the ones obtained for each hemisphere without combining the two decision spaces together. For all comparisons, the performance was improved significantly after applying the global classifiers. Moreover, as shown in Table \ref{table3}, for the most difficult two groups to delineate, CN vs. MCI, the global model yielded significant improvement of all classification statistics comparing to those recently reported cross validation results [14]–[16], that the proposed study achieves the best performance.  

\subsection{Classification performance on held-out test data}

In order to obtain a reliable measure of the proposed CAD system performance, the remaining $20\%$ of the noise-free detected data points were used as the held-out test data applying the acquired optimal sets of features that results are presented in Table \ref{table4}, which achieved significant performance improvement even in contrast to those recent reported cross validation results in Table \ref{table3}, especially, for discriminating the most challenging MCI group from CN. 

\begin{table}[t]
	\begin{adjustwidth}{-1in}{-1in}
	\caption{Summary of tenfold cross validation performance for each model}
	\label{table2}
	\centering
	\begin{tabular}{lllllllllllll}
		\toprule
		Comparison    
		& \multicolumn{4}{c}{CN vs. MCI}     
		& \multicolumn{4}{c}{CN vs. AD}     
		& \multicolumn{4}{c}{MCI vs. AD}  \\
		\cmidrule{2-13}
		Feature ranking     
		& \multicolumn{2}{c}{Local}  & \multicolumn{2}{c}{Global}  
		& \multicolumn{2}{c}{Local}  & \multicolumn{2}{c}{Global}  
		& \multicolumn{2}{c}{Local}  & \multicolumn{2}{c}{Global}  \\
		\cmidrule{2-13}
		Classifier    
		& Loc.  & Glo.   & Loc.   & Glo.    
		& Loc.  & Glo.   & Loc.   & Glo.    
		& Loc.  & Glo.   & Loc.   & Glo.  \\
		\midrule
		
		F1 score$\%$	
		& 88.11 & 92.08 & 83.62 & 95.30 & 92.66 & 95.89 & 91.67 & 96.23 & 74.69 & 81.41 & 69.14 & 78.49 \\
		Accuracy$\%$	
		& 83.36 & 90.26 & 75.90 & 94.10 & 93.85 & 96.54 & 93.08 & 96.92 & 82.57 & 89.43 & 78.57 & 88.57 \\
		Sensitivity$\%$	
		& 94.17 & 92.08 & 100.0 & 97.08 & 91.82 & 95.45 & 90.00 & 92.73 & 81.82 & 73.64 & 76.36 & 66.36 \\
		Specificity$\%$	
		& 68.67 & 87.33 & 37.33 & 89.33 & 95.33 & 97.33 & 95.33 & 100.0 & 82.92 & 96.67 & 79.58 & 98.75 \\
		\midrule
		Side of the brain
		& \multicolumn{12}{c}{Optimal Feature Sets} \\
		\cmidrule{2-13}
		Left 		& 36 & 5  & 64 & 69 & 6  & 10 & 6  & 4  & 2  & 48 & 1  & 1  \\
		Right 		& 34 & 44 & 65 & 1  & 5  & 44 & 5  & 42 & 2  & 4  & 2  & 61 \\
		\bottomrule
	\end{tabular}
	\end{adjustwidth}
\end{table}

\begin{table}[t]
	\begin{adjustwidth}{-1in}{-1in}  
	\caption{Summary of tenfold cross validation performance for each model}
	\label{table3}
	\centering
	\begin{tabular}{llllllll}
		\toprule
		\multicolumn{2}{c}{Comparison}    
		& \multicolumn{3}{c}{CN vs. MCI}     
		& \multicolumn{3}{c}{CN vs. AD}    \\
		\cmidrule{3-8}
		Reference	& Modalities	
		& ACC $\%$	& SEN $\%$	& SPE $\%$	 
		& ACC $\%$	& SEN $\%$	& SPE $\%$ \\
		\midrule
		Khedher L., et al. (2015) [14]	& MRI
		& 81.89	& 82.61	& 81.62
		& 88.49	& 91.27	& 85.11				\\
		Tong T.,	et al. (2017) [15]		& MRI+PET+CSF+Genetic
		& 79.50	& 85.10	& 67.10
		& 91.80	& 88.90	& 94.70				\\
		Khedher L., et al. (2017) [16]	& MRI
		& 79.00	& 82.00	& 76.00
		& 89.00	& 92.00	& 86.00				\\
		Proposed Study	& MRI
		& \textbf{94.10}	& \textbf{97.08}	& \textbf{89.33}
		& \textbf{96.92}	& \textbf{92.73}	& \textbf{100.0}	\\
		\bottomrule
	\end{tabular}
	\end{adjustwidth}
\end{table}

\begin{table}[b]
	\caption{Summary of the classification performance using held-out test data}
	\label{table4}
	\centering
	\begin{tabular}{lllll}
		\toprule
		Comparison    			 
		& F1 score $\%$ & Accuracy$\%$	& Sensitivity$\%$	& Specificity$\%$ \\
		\cmidrule{2-5}
		CN vs. MCI
		& \textbf{91.06}	& \textbf{88.78}	& \textbf{91.80}	& \textbf{83.78}	\\
		CN vs. AD
		& \textbf{90.48}	& \textbf{93.10}	& \textbf{90.48}	& \textbf{94.59}	\\
		\bottomrule
	\end{tabular}
\end{table}

\subsubsection*{Acknowledgments}
We acknowledge the critical support provided by the National Science Foundation under grants: CNS-1532061, CNS-1551221, CNS-1642193, IIP 1338922, and CNS-1429345. The generous support of the Ware Foundation is greatly appreciated. This research is also supported through the Florida Department of Health, Ed and Ethel Moore Alzheimer's Disease Research Program, and Florida ADRC (Alzheimer’s Disease Research Center) (1P50AG047266-01A1).
 
Data used in preparation of this article were obtained from the Alzheimer’s Disease Neuroimaging Initiative (ADNI) database (adni.loni.usc.edu). As such, the investigators within the ADNI contributed to the design and implementation of ADNI and/or provided data but did not participate in analysis or writing of this report. A complete listing of ADNI investigators can be found at: \url{http://adni.loni.usc.edu/wp-content/uploads/how\_to\_apply/ADNI\_Acknowledgement\_List.pdf}. Data collection and sharing for this project was funded by the Alzheimer's Disease Neuroimaging Initiative (ADNI) (National Institutes of Health Grant U01 AG024904) and DOD ADNI (Department of Defense award number W81XWH-12-2-0012).
%

\section*{References}

\small

[1] Cuingnet, R., Gerardin, E., Tessieras, J., Auzias, G.\ et al. (2011) Automatic classification of patients with Alzheimer’s disease from structural MRI: A comparison of ten methods using the ADNI database. {\it Neuroimage}, {\bf 56}(2):766–781. 

[2] Fang, C., Li, C., Cabrerizo, M., Barreto, A.\ et al. (2017) A Novel Gaussian Discriminant Analysis-based Computer Aided Diagnosis System for Screening Different Stages of Alzheimer’s Disease. {\it Proceedings of 17th IEEE BIBE}:279–284.

[3] Braak, H.\  \& Tredici, K.D.\ (2012) Alzheimer’s disease: pathogenesis and prevention. {\it Alzheimers  Dement.}, {\bf 8}(3):227–233. 

[4] Gómez-Isla, T., Price, J.L., McKeel, D.W., Morris, J.C., Growdon, J.H.\ \& Hyman, B.T.\ (1996) Profound loss of layer II entorhinal cortex neurons occurs in very mild Alzheimer's disease. {\it J. Neurosci.}, {\bf 16}(14):4491–4500. 

[5] Khan, U.A., Liu, L., Provenzano, F.A., Berman, D.E., Profaci, C.P., Sloan, R., Mayeux, R., Duff, K.E.\ \& Small, S.A.\ (2013) Molecular drivers and cortical spread of lateral entorhinal cortex dysfunction in preclinical Alzheimers disease. {\it Nat. Neurosci.}, {\bf 17}(2):304–311. 

[6] Bakkour, A., Morris, J.C.\ \& Dickerson B.C.\ (2009) The cortical signature of prodromal AD: regional thinning predicts mild AD dementia. {\it Neurology}, {\bf 72}:1048-1055. 

[7] University of South California,\ ADNI |  Standardized MRI Data Sets. ADNI. [Online]. Available: \url{http://adni.loni.usc.edu/methods/mri-analysis/adni-standardized-data/} [Accessed: 02-Aug-2017]. 

[8] Reuter, M., Schmansky, N.J., Rosas, H.D.\ \& Fischl, B.\ (2012) Within-subject template estimation for unbiased longitudinal image analysis. {\it Neuroimage}, {\bf 61}(4):1402–1418.
 
[9]	Tustison, N.J., Cook, P.A., Klein, A., Song, G., Das, S.R., Duda, J.T., Kandel, B.M., Strien, N.V., Stone, J.R., Gee, J.C.\ \& Avants, B.B.\ (2014) Large-scale evaluation of ANTs and FreeSurfer cortical thickness measurements. {\it Neuroimage}, {\bf 99}:166–179.

[10] Klein, A., Ghosh, S.S., Bao, F.S., Giard, J., Hame, Y., Stavsky, E., Lee, N., Rossa, B., Reuter, M., Neto, E.C.\ \& Keshavan, A.\ (2017) Mindboggling morphometry of human brains. {\it PLoS Comput. Biol.}, {\bf 13}(2):e1005350. 

[11] Alzheimer's Disease Big Data DREAM Challenge 1. [Online]. Available: \url{https://www.synapse.org/#!Synapse:syn2290704/wiki/60828} [Accessed: 02-Aug-2017]. 

[12] Mindboggle Data. Mindboggle-101. [Online]. Available:\url{http://www.mindboggle.info/data.html} [Accessed: 02-Aug-2017]. 

[13] Squire, L.R.\ \& Zola-Morgan, S.\ (1991) The medial temporal lobe memory system. {\it Science}, {\bf 253}(5026):1380–1386. 

[14] Khedher, L., Ramírez, J., Górriz, J.M., Brahim, A., Segovia, F.\ et al. (2015) Early diagnosis of Alzheimer’s disease based on partial least squares, principal component analysis and support vector machine using segmented MRI images. {\it Neurocomputing}, {\bf 151}(1):139–150. 

[15] Tong, T., Gray, K., Gao, Q., Chen, L.\ \& Rueckert, D.\ (2017) Multi-modal classification of Alzheimer's disease using nonlinear graph fusion. {\it Pattern Recognition}, {\bf 63}:171–181. 

[16] Khedher, L., Illán, I., Górriz, J.M., Ramírez, J., Brahim, A.\ \& Meyer-Baese, A.\ (2017) Independent Component Analysis-Support Vector Machine-Based Computer-Aided Diagnosis System for Alzheimer's with Visual Support. {\it Int. J. Neural Syst.}, {\bf 27}(3):1650050. 

\end{document}